\documentclass[letterpaper, 10 pt, conference]{ieeeconf}  %

\IEEEoverridecommandlockouts                              %

\usepackage{amssymb}  %
\usepackage{svg}
\usepackage{pifont}
\usepackage{multirow}
\usepackage{flushend}

\usepackage{graphics}           
\usepackage{times}              
\usepackage{amsmath}            
\usepackage{amssymb}            
\usepackage{graphicx}
\usepackage{algorithm}
\usepackage[noend]{algpseudocode}
\usepackage{booktabs}
\usepackage{color}
\usepackage{listings}
\usepackage{subfiles}
\usepackage{hyperref}
\usepackage[nocompress]{cite} %
\definecolor{instructioncolor}{rgb}{.5,.5,.5}

\usepackage[font=small]{caption}

\def\secref#1{Sec.~\ref{#1}}
\def\figref#1{Fig.~\ref{#1}}
\def\tabref#1{Tab.~\ref{#1}}
\def\eqref#1{Eq.~(\ref{#1})}

\makeatletter
\usepackage{xspace}
\DeclareRobustCommand\onedot{\futurelet\@let@token\@onedot}
\def\@onedot{\ifx\@let@token.\else.\null\fi\xspace}
\def\eg{e.g\onedot} 
\def\ie{i.e\onedot}

\def\etal{{et al}\onedot}
\makeatother

\usepackage{array}
\newcolumntype{L}[1]{>{\raggedright\let\newline\\\arraybackslash\hspace{0pt}}m{#1}}
\newcolumntype{C}[1]{>{\centering\let\newline\\\arraybackslash\hspace{0pt}}m{#1}}
\newcolumntype{R}[1]{>{\raggedleft\let\newline\\\arraybackslash\hspace{0pt}}m{#1}}

\renewcommand{\b}[1]{\mbox{\boldmath$#1$}}

\renewcommand{\v}[1]{{\b #1}}

\newcommand{\name}{Score}
\newcommand{\yolo}{YOLOv9}
\newcommand{\bev}{bird's eye view}
\newcommand{\openlane}{OpenLane-V2}
\newcommand{\argo}{Argoverse 2}
\newcommand{\nuscenes}{nuScenes}

\newcolumntype{C}[1]{>{\centering\arraybackslash}p{#1}}
\newcommand{\cmark}{\ding{51}}%
\newcommand{\xmark}{\ding{55}}%
\title{\LARGE \bf Coherent Online Road Topology Estimation and Reasoning \\with Standard-Definition Maps}

\author{Khanh Son Pham* \and Christian Witte* \and Jens Behley \and Johannes Betz \and Cyrill Stachniss%
\thanks{* Equal contribution. S. Pham is with CARIAD SE and the Technical University Munich, Germany.
C. Witte is with CARIAD SE and with the Center for Robotics, University of Bonn, Germany.
J. Betz is with the Technical University Munich, Germany and the Munich Institute of Robotics and Machine Intelligence~(MIRMI), Germany.
J. Behley, C. Stachniss are with the Center for Robotics, University of Bonn, Germany. C. Stachniss is also with the Lamarr Institute for Machine Learning and Artificial Intelligence, Germany.
}
}

\begin{document}

\maketitle
\thispagestyle{empty}
\pagestyle{empty}

\begin{abstract}

  Most autonomous cars rely on the availability of high-definition (HD) maps.
  Current research aims to address this constraint by directly predicting HD map elements from onboard sensors and reasoning about the relationships between the predicted map and traffic elements.
  Despite recent advancements, the coherent online construction of HD maps remains a challenging endeavor, as it necessitates modeling the high complexity of road topologies in a unified and consistent manner. 
  To address this challenge, we propose a coherent approach to predict lane segments and their corresponding topology, as well as road boundaries, all by leveraging prior map information represented by commonly available standard-definition (SD) maps.  
  We propose a network architecture, which leverages hybrid lane segment encodings comprising prior information and denoising techniques to enhance training stability and performance. Furthermore, we facilitate past frames for temporal consistency.
  Our experimental evaluation demonstrates that our approach outperforms previous methods by a significant margin, highlighting the benefits of our modeling scheme.  
\end{abstract}

\section{INTRODUCTION}
\label{sec:intro}

As a prevailing robotic application, the operation of autonomous vehicles in urban environments is contingent upon the availability of high-definition maps of the surrounding environment.
These HD maps are supposed to contain precise geometric information about the drivable lanes, including their correspondences with one another, and their relationship to other important traffic elements, such as traffic lights and signs. The generation of such HD maps is a time-consuming and resource-intensive semi-automated process that involves manual steps. At the same time, these maps are subject to changes over time~\cite{jiao2018compsac}.
For safe and robust autonomous driving, the detection of traffic elements such as traffic lights and traffic signs and their association with lanes or lane segments is crucial. For example, traffic light states or prohibitive traffic signs contain important information for predicting the future trajectories of other traffic participants.

Recent approaches \cite{liu2023icml, yuan2024wacv} have demonstrated encouraging results in the detection of HD map elements. However, they do not achieve the objective of computing consistent and accurate HD maps online while driving. The latest approaches reformulate map element detection into a lane segment formulation~\cite{li2024iclr} to leverage the known geometric properties of lane layouts and further predict the topological relationship of lane segments or centerlines~\cite{li2023arxiv-gtrf}. 
Other lines of work~\cite{luo2024icra} seek to integrate prior knowledge about road geometry into their architectural design by incorporating standard-definition map data into the \bev\ (BEV) transformation.

None of the these approaches have adequately addressed the challenge of jointly detecting lane segments, their topological relationships, while exploiting available a-priori knowledge in the form of SD maps. 
For this reason, the CVPR 2024 Grand Autonomous Challenge, with its \textit{Mapless Driving} track, proposed a problem formulation for solving these tasks simultaneously. 

\begin{figure}[tb]
  \centering
  \includegraphics[height=5.7cm]{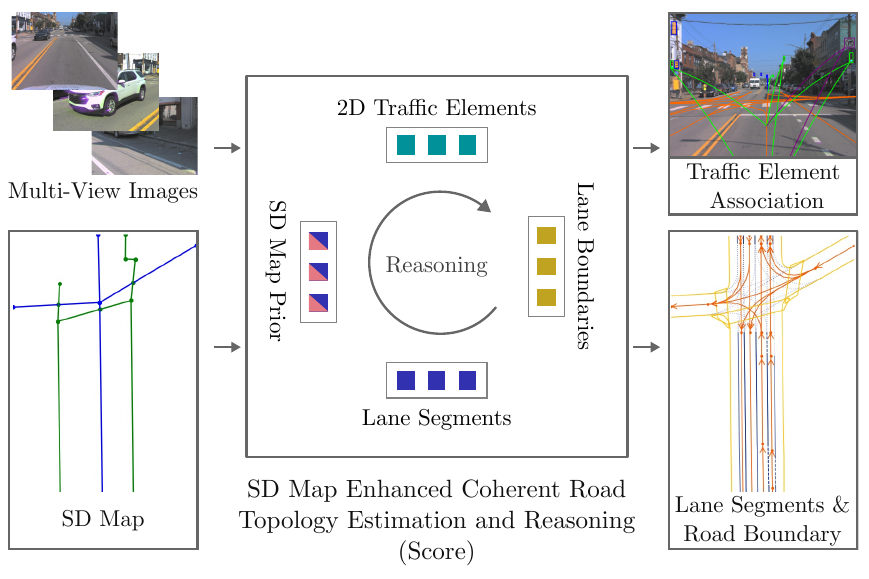}
  \caption{Given multi-view images and a standard-definition (SD) map, our method predicts lane segments and lane boundaries, as well as traffic elements. It reasons about the topology among lane segments and estimates the traffic element-to-lane association.}
  \label{fig:motivation}
\end{figure}

To this end, we approach to solve the union of tasks coherently, that is, to consider them as a whole with clear interfaces.
Thus, we systematically design a network architecture to predict lane segments and their topological relationships by leveraging latent embeddings (queries), as depicted in \figref{fig:motivation}. It encompasses the topology among lane segments, as well as the topology between lane segments and traffic elements (e.g., traffic lights or traffic signs).
Given an SD map, we ingest prior information as enhanced hybrid queries with a positional sampling module. This provides the network hypotheses of potential lane segments, which can be iteratively validated and refined by interacting with the BEV features. We introduce temporal fusion and lane denoising for more consistent and robust predictions, while we optimize the overall performance by suggesting further improvements to the architecture.
The experimental results demonstrate the efficacy of the proposed method, which reaches state-of-the-art performance on the \openlane~\cite{wang2023neurips} dataset.

More qualitative results and additional visualizations can be found on our project page: \url{https://www.ipb.uni-bonn.de/html/projects/score}.

\section{RELATED WORK}

To mitigate the reliance on expensive high-definition maps, recent works have focused on deriving vectorized representation of HD map elements as polylines.
This representation is commonly predicted in \bev\ (BEV) space from onboard sensors.

\textbf{Bird's Eye View Transformation}.
Using a categorical depth distribution, Lift-Splat-Shoot (LSS)~\cite{philion2020eccv} and their derivative works~\cite{huang2021arxiv-bevdet, li2023aaai, liu2023icra, zhou2023iccv} project features onto a 3D volume and then collapse the volume onto a 2D plane with efficient pooling. Transformer-based approaches~\cite{li2022eccv, yang2023cvpr, liu2022eccv, liu2023iccv, zhou2022cvpr, chen2022arxiv, witte2025wacv} cross-attend BEV queries to images features and thus extract a BEV representation. 

\textbf{HD Map Generation}.
Leveraging semantic segmentation, HDMapNet~\cite{li2022icra-haoh} utilizes a post-processing step to output a polyline representation of map elements. In a two stage approach, VectorMapNet~\cite{liu2023icml} first identifies map elements and then outputs polylines using a detection transformer~\cite{carion2020eccv}. MapTR~\cite{liao2023iclr} proposes to directly predict polylines in a single-stage approach by employing hierarchical queries, while the authors introduce one-to-many assignment, decoupled attention, and auxiliary losses in their follow-up work~\cite{liao2024ijcv}. Further improvements to MapTR include predicting pivotal points~\cite{ding2023cvpr}, modeling the output as Bézier curves~\cite{qiao2023cvpr}, incorporating geometric properties into the learning process~\cite{zhang2024eccv, yu2023arxiv-ssml}.
Another line of work puts emphasis on the query generation, e.g., by leveraging instance~\cite{liu2024cvpr} or semantic~\cite{choi2024eccv} segmentation masks or by introducing hybrid queries~\cite{xu2024eccv, zhou2024cvpr}.

\textbf{Topology Estimation}. 
Understanding road topology is crucial for scene comprehension in autonomous driving, which includes predicting lane centerlines and their topology, \ie, constructing a lane graph with successor-predecessor relationship.
Can \etal~\cite{can2021cvpr} perform the graph estimation for single monocular images by associating and merging piecewise centerline estimates based on a predictions of a transformer-based model. 
Building on this, TopoRoad~\cite{xu2023icra} introduces minimal cycle queries to maintain the correct order of intersections. LaneGAP~\cite{liao2024eccv} uses a heuristic algorithm to reconstruct the graph from a set of lanes. 
TopoNet~\cite{li2023arxiv-gtrf} transforms 2D detections for traffic elements into a unified feature space to model the relationship with a scene graph neural network, while in TopoMLP~\cite{wu2024iclr}, the authors emphasis the importance of a strong 2D detection performance.
Other works combine trajectory knowledge and visual SLAM to estimate the lane information~\cite{opra2024iros} or extend the topology estimation idea by leveraging geometric properties \cite{fu2024neurips} or queries initialized by 2D priors \cite{li2024arxiv-e3ld}.
Li \etal~\cite{li2024iclr} propose to predict lane segments, each consisting of semantic meaningful lane entities defined by a lane centerline, lane boundaries, as well as successor relationship and class information. 

\textbf{SD Map-aided Map Generation}.
To improve upon the lane-lane relationship estimation, SMERF~\cite{luo2024icra} integrates prior information in form of standard-definition maps into the network architecture by encoding the SD map elements with a transformer encoder and by ingesting the feature representation into the BEV transformation process. 
Other approaches propose to fuse the SD map information in a rasterized fashion~\cite{ma2024eccv} or to ingest the knowledge at different stages of the network~\cite{zhang2024iros, yang2024arxiv-ttls}.

\textbf{Temporal Fusion}.
To achieve temporal consistency, previous works~\cite{li2022eccv, huang2022arxiv, han2024ral} have shown various methods on fusing past knowledge with the current observations. One line of work cross-attends to image features of previous timesteps by back-projecting reference points and aggregating features for semantic segmentation or 3D object detection~\cite{li2022eccv, lin2022arxiv}, while other methods~\cite{huang2022arxiv, yang2023cvpr} stack the current BEV features with motion-compensated BEV features from the previous timestep.
Further work recurrently fuses BEV features~\cite{han2024ral} or propagates high confidence queries to the next timestep~\cite{lin2023arxiv-svae}.
StreamMapNet~\cite{yuan2024wacv} introduces the idea of streaming temporal fusion to HD map element detection by recurrently fusing BEV features and propagating past queries. Wang \etal~\cite{wang2024eccv-sqdf} also apply denoising to enhance lane queries for temporal fusion.

\textbf{Mapless Driving}. The CVPR 2024 Grand Autonomous Challenge introduced the joint tasks of lane segment detection, road boundary estimation, and traffic element association with SD map information.
Participating works~\cite{yang2024arxiv, wu2024arxiv} address this joint formulation by integrating the prior SD map information similar to \cite{luo2024icra}, and use large backbones and ensembles to improve performance.

\begin{figure*}[tb]
  \centering
  \framebox{\includegraphics[height=8.5cm]{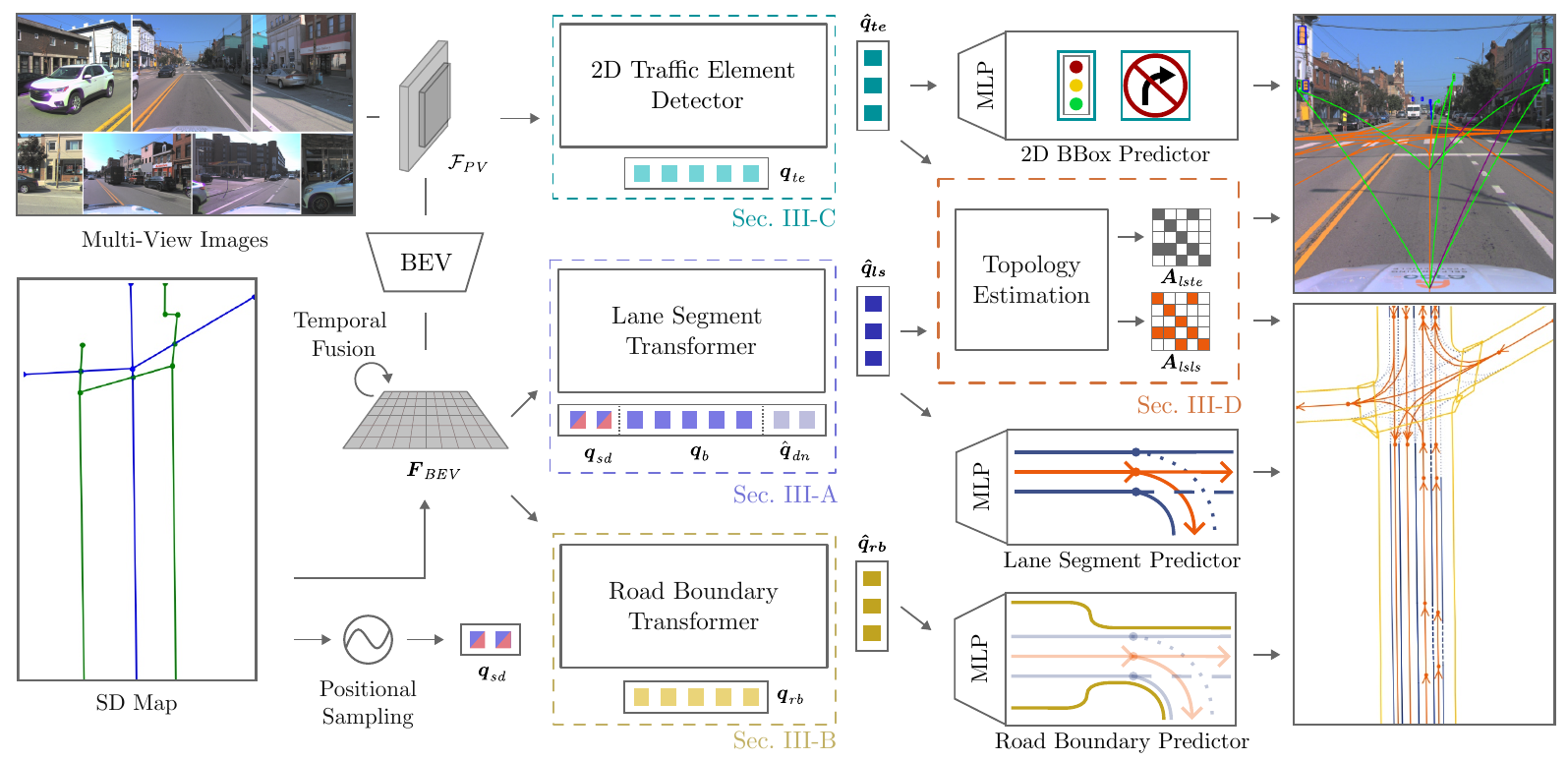}}
  \vspace{-0.1cm}
  \caption{Overall architecture of our approach. First, multi-view image features are transformed into a BEV representation. Utilizing SD map prior information, we enhance learnable queries by a positional sampling and add lane denoising queries. In the \textit{Lane Segment Transformer}, lane segment queries (blue) are refined with BEV features, while two further transformer decoder output refined road boundary queries (yellow) and 2D traffic element queries (cyan). In the \textit{Topology Estimation} module, these queries interact to derive the lane-to-lane and traffic element-to-lane association. Simple MLPs are employed to decode the queries $\hat{\v{q}}$ into the corresponding output format.}
  \label{fig:architecture}
  \vspace{-0.2cm}
\end{figure*}

\section{SD Map Enhanced Coherent Road Topology Estimation and Reasoning}

We aim to predict HD map elements and their corresponding relationships based on onboard multi-view images. To this end, we propose SD map enhanced coherent road topology estimation and reasoning, called \textbf{\name}, which jointly predicts lane segments, road boundaries, and traffic elements such as traffic lights or traffic signs. Further, our method derives an association between the traffic elements and the lane segments, as well as the topology among the lane segments. 

Given multi-view images, the image backbone outputs multi-scale image feature maps $\mathcal{F}_{\textit{PV}} = \{ \v{F}_{\text{PV}}^i\} $ for the $i$-th camera, while top-view features $\v{F}_{\text{BEV}}$ are obtained by a BEV transformation~\cite{li2022eccv}. To leverage the geometric constraints of road lanes, we utilize the lane segment formulation~\cite{wang2023neurips}. A lane segment ($\text{ls}$) is defined as a set of polylines, representing the centerline and lane boundaries, along with a classification label indicating the type of lane segment and lane boundary types (\eg, solid, dashed, or invisible). Further, we consider pedestrian crossing as lane segments.

As a separate task, we infer road boundaries ($\text{rb}$) as polylines with $n_\textit{rb}$ points, while the traffic elements ($\text{te}$) are predicted in 2D with bounding box annotation.
We denote the topology among lane segments as lane graph $\left( \v{V}_{\text{ls}}, \v{A}_{\text{lsls}} \right)$ with $\v{V}_{\text{ls}}$ as the set of lane segments and the edge set $\v{A}_{\text{lsls}}$ as the adjacency matrix~\cite{li2023arxiv-gtrf, li2024iclr}. Similarly, we denote the correspondence between lane segment and traffic elements as a bipartite graph $\left( \v{V}_{\text{ls}} \cup \v{V}_{\text{te}}, \v{A}_{\text{lste}} \right)$. 
The SD map is provided as a set of polylines with class labels (\eg, road, sidewalk).

An overview of our proposed architecture \name\ is illustrated in \figref{fig:architecture}.
The network is composed of multiple components, which utilize queries as interfaces.
The lane segment transformer (\secref{sec:lanehead}) and road boundary transformer (\secref{sec:roadhead}) use the BEV features to output instance-level map element embeddings.
The 2D traffic element detector outputs traffic sign and traffic light detections based on 2D image features (\secref{sec:2dhead}). In the topology estimation (\secref{sec:topohead}), the network derives adjacencies for the lane-lane and lane-traffic element topology.

\subsection{Lane Segment Transformer}
\label{sec:lanehead}
To infer lane segments, our model utilizes a similar architecture to LaneSegNet~\cite{li2024iclr}, incorporating a transformer-based decoder. 
In the transformer decoder, the hybrid queries~$\v{q_\textit{ls}}$ cross-attend to the enriched BEV features $\v{F}_{\text{BEV}}$, which are then refined iteratively using self-attention~\cite{vaswani2017nips} and deformable attention~\cite{xia2022cvpr}.
To address the limitations of deformable attention in capturing long-range dependencies, particularly for elongated lane shapes, we integrate lane attention, a lane segment-aware deformable attention~\cite{li2024iclr}. This module distributes multiple reference points along the estimated lane boundaries of the previous transformer decoder layer. 
We enhance the lane decoding by leveraging hybrid queries and condition the decoding on the prior knowledge in form of reference points. Then, the refined queries~$\hat{\v{q}}_\textit{ls}$ are calculated as:
\begin{align}
\hat{\v{q}}_{\textit{ls}} = \textrm{LaneDecoder}\left(\v{F}_{\textit{BEV}}, \v{q}_{\textit{ls}} \mid \v{p}_{\textit{ls}}\right) \in \mathbb{R}^{N_{\textit{ls}}\times C},
\end{align}
with $\v{q}_\textit{ls}=\mathrm{concat}(\v{q}_{\textit{sd}}, \v{q}_{b}, \hat{\v{q}}_{\textit{dn}})$ comprise of SD-enhanced queries $\v{q}_{\textit{sd}}$, denoising queries $\hat{\v{q}}_{\textit{dn}}$, both of which will be explained in the following sections, and the base queries $\v{q}_{b}$. $\v{p}_{\textit{ls}}=\mathrm{concat}(\hat{\v{p}}_{\textit{sd}}, \v{p}_{b}, \hat{\v{p}}_{\textit{dn}})$ denote their respective reference points and $C$ is the channel dimension.

\subsubsection{SD Map Enhanced Queries}

Initially, for each query, a randomly initialized reference point is generated, as in LaneSegNet~\cite{li2024iclr}, which has no prior driving scene information. To improve the initial placement, we propose incorporating geometric data from the SD map, thus, leveraging prior knowledge as potential lane segment hypotheses. 
For that, we define the set of $n_\textit{sd}$ queries $\v{q}_\textit{sd} \in \mathbb{R}^{n_\textit{sd} \times C}$ and generate reference points for these queries by sampling one point from the middle of each edge and continue by sampling more points proportionally based on edge length until we reach $n_{\textit{sd}}$ points. This ensures evenly-spaced sampling and prevents duplicative information propagation.
For the $i$-th SD map query, the initial reference point placement is computed as follows:
\begin{align}
    \hat{\v{p}}_\textit{sd,i} = \text{MLP}(\v{q}_\textit{pos,i}) + \v{p}_\textit{sd,i},
    \label{eq:sdrefpoints}
\end{align}
where $\v{q}_\textit{pos,i}$ denotes the positional encoding of the $i$-th query and  $\v{p}_\textit{sd,i}$ depicts the sampled reference points, which are normalized to $[0, 1]$.
This formulation allows for both, a learnable and an a-priori term, providing the network further flexibility.
Lastly, a self-attention mask between SD map enhanced queries $\v{q}_\textit{sd}$ and other queries prevents the leakage of deficient information, as SD map can have annotation errors such as missing or wrong map elements.

\subsubsection{Lane Denoising}
 
To improve convergence of detection transformer (DETR)-style networks, Li \etal~\cite{li2022cvpr-dadt} propose the concept of query denoising. The key idea is to employ ground truth queries augmented with noise to the training process, which effectively addresses the inherent instabilities associated with bipartite matching~\cite{li2022cvpr-dadt}.
Inspired by this approach, we introduce lane denoising. 
For $ i \in \{1, ..., N_{\textit{GT}}\}$, random noise $\Delta \v{p}_\textit{dn,i} = (\Delta x, \Delta y, \Delta z) \overset{\text{i.i.d.}}{\sim}~\mathcal{U}(-1, 1)$ is added to the $N_{\textit{GT}}$ ground truth annotations, comprising three polylines with $n_{\textit{pts}}$ points each, representing the centerline and the left and right boundaries. Along with the their reference points $\hat{\v{p}}_\textit{dn} = \lambda_\textit{dn} \left( \v{p}_\textit{dn} +  \Delta \v{p}_\textit{dn}\right)$, the final denoising queries are obtained as:
\begin{align}
    \hat{\v{q}}_\textit{dn} = \v{q}_\textit{dn} + \text{MLP}(\hat{\v{p}}_{\textit{dn}}),
    \label{eq:denoising}
\end{align}
with $\lambda_\textit{dn} \in (0, 1)$ being a hyperparameter for rescaling. Similar to DN-DETR~\cite{li2022cvpr-dadt}, we introduce lane denoising groups, each consisting of a noisy version of all ground truth lane segments. The number of denoising groups is computed as $\lfloor\frac{n_\textit{dn}}{N_\textit{GT}}\rfloor$. For each group, we derive a different set of initial reference points from the noisy ground truth to prevent propagating similar information. Finally, to avoid ground truth information leakage~\cite{li2022cvpr-dadt}, masked attention is used to prevent the denoising groups from interacting with other queries.

\subsection{Road Boundary Transformer}
\label{sec:roadhead}

In contrast to the lane segment prediction, the detection of road boundaries necessitates the inference of irregularly shaped polylines that differentiate between drivable and non-drivable area. 
Thus, we reformulate our lane segment head to accommodate this task to model the road boundary instances as queries and finally output polylines. As opposed to prior work, \eg, MapTR~\cite{liao2023iclr}, we do not employ hierarchical queries, but leverage one query per instance reducing the computational requirements. 
Inspired by multi-point attention~\cite{yuan2024wacv}, we predict $n_{\textit{rb}}$ points for the road boundary for each intermediate transformer decoder layer, which are used in the consecutive decoder layer to sample $n_{\textit{ref}}$ equidistant reference points.  

\subsection{2D Traffic Element Detection}
\label{sec:2dhead}
To associate traffic elements with lane segments, we first detect traffic lights and traffic signs as 2D bounding boxes using the image features of the front camera.

Given our query-based approach, we utilize DN-DETR~\cite{li2022cvpr-dadt}, a denoising Deformable DETR for 2D bounding box detection. Traffic element queries $\v{q}_{\textit{te}}$ interact with the image features~$\v{F}_{\textit{PV}}^0$ of front camera and the traffic decoder outputs the most confident predictions as queries. 
Similar to TopoMLP~\cite{wu2024iclr}, we assess the usage of proposals generated by \yolo~\cite{wang2024eccv} and incorporate them as additional queries. For simplicity, we denote all traffic element queries as $\v{q}_{te}\in \mathbb{R}^{N_{\textit{te}}\times C}$. 
Thus, the refined queries for traffic elements are computed as
\begin{align}
\hat{\v{q}}_{\textit{te}} = \textrm{TrafficDecoder}(\v{F}_{\textit{PV}}^0, \v{q}_{\textit{te}}).
\end{align}

Further, we employ \yolo~\cite{wang2024eccv} and propose to encode the bounding box parameters (\eg, center, height, etc.) into the embedding space with a simple MLP to obtain $\hat{\v{q}}_{\textit{te}}$.

\begin{figure*}[tb]
  \centering
  \framebox{\includegraphics[height=5.0cm]{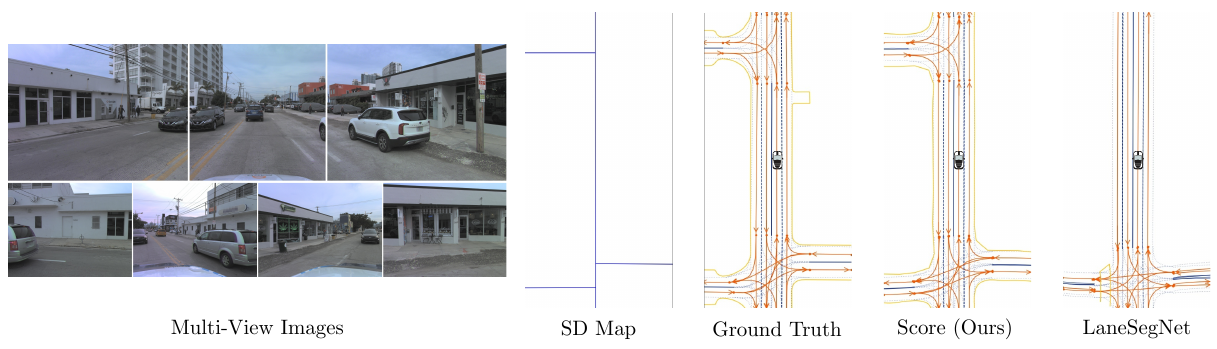}}
  \vspace{-0.2cm}
  \caption{Qualitative results. Our method demonstrates the capacity to incorporate a SD map prior and past information in an effective manner, facilitating the prediction of partly-occluded roads and providing more consistent predictions than those produced by LaneSegNet~\cite{li2024iclr}.}
  \label{fig:qualitative result}
  \vspace{-0.4cm}
\end{figure*}

\subsection{Topology Estimation}
\label{sec:topohead}

The topology estimation is designed to facilitate reasoning about the pairwise relationships between lane instances within the same embedding space~\cite{can2021cvpr, li2023arxiv-gtrf}, as well as to associate traffic elements with lane segments. 
Given the lane segments queries $\hat{\v{q}}_{\textit{ls}}$ and traffic elements queries $\hat{\v{q}}_{\textit{te}}$, our topology head encodes the queries and repeats the lane segment queries $N_{\textit{te}}$ times and the traffic element queries $N_{\textit{ls}}$ times. The resulting matrices are stacked such that we obtain $\v{Q}_{\textit{lste}} \in \mathbb{R}^{N_{\textit{ls}} \times N_{\textit{te}} \times 2C}$.
Similar to~\cite{wu2024iclr, li2024iclr}, we compute the similarity score $\v{A}^S_{\textit{lste}} \in [0, 1]^{N_{\textit{ls}} \times N_{\textit{te}}}$ between the instance pair $(i, j)$ as an adjacency matrix:
\begin{align}
    A_{\textit{lste}}^S(i,j) = \sigma\left( \text{MLP}(\v{Q}_{\textit{lste}}(i, j)) \right),
\end{align}
with $\sigma$ being the sigmoid function. The same operation can be applied straight-forward to estimate the topology $\v{A}_{\textit{lsls}}^S$ among all $\v{q}_\textit{ls}$ instances, \ie, lane segments.

Inspired by Topologic \cite{fu2024neurips}, we design a post-processing function that maps the absolute distance between the start and end points of all possible lane instance pairs to a distance-based topology matrix $\v{A}_\textit{lsls}^D\in [0, 1]^{N_{\textit{ls}\times} N_{\textit{ls}}}$: 
\begin{align}
A_{\textit{lsls}}^D(i,j) = \frac{2}{1 + \exp{\left(\frac{d_{i,j}}{\alpha}\right)}},
\end{align}
with $d_{i, j}$ being the distance between the start/end point of lane segment $i$ and $j$ and $\alpha\in\mathbb{R}$ denoting a parameter for the distance matching.
The final relationship score ${A}_{\textit{lsls}}$ is calculated as follows:
\begin{equation}
{A}_{\textit{lsls}}(i,j) = \min\left(A_{\textit{lsls}}^S(i, j) +  \beta A_{\textit{lsls}}^D(i, j),\ 1\right),
\end{equation}
where $\beta$ denotes a weight factor to account for the distance-based similarity. 
Through empirical analysis, we determined that the best performance was achieved with $\alpha = 2.5$ and $\beta = 0.8$.

\subsection{Holistic Improvements}
\label{sec:improvements}
We introduce further improvements to the query-based lane segment transformer and road boundary transformer with the aim of enhancing the detection performance.
\subsubsection{SMERF}
We employ the idea of SMERF~\cite{luo2024icra} and cross-attend the encoded SD map with the BEV features.
\subsubsection{One-to-Many (o2m)}
To further accelerate convergence and improve detection accuracy, we integrate one-to-many matching~\cite{jia2023cvpr} for lane segment and road boundary detection.
\subsubsection{Dataset Resampling}
An analysis of the distribution of categories within the dataset reveals a significant imbalance that can have implications for the model's performance. Specifically, statistical analysis shows that nearly 96\% of the training samples involve the vehicle moving straight, while only 4\% contain to turning maneuvers. 
To address this issue, we increase the number of samples in scenarios where the wheel angle is greater than 7°.
\subsubsection{Temporal Fusion}
To leverage also past observations, we employ temporal fusion analogous to StreamMapNet~\cite{yuan2024wacv}. In a streaming fashion, the BEV features $\v{F}_{\textit{BEV}}$ are recurrently fused with the previous BEV features utilizing a ConvGRU \cite{chung2014nipsws} to integrate temporal information. Subsequently, the fused BEV features undergo ego-motion compensation and are being propagated to the next timestep. We initially conduct a warm-up phase~\cite{yuan2024wacv, chen2024eccv}, \ie, an initial training that does not involve past frames, and then subsequently employ temporal fusion.

\subsection{Loss Functions}
\label{sec:losses}
Our model employs the Hungarian algorithm to match predictions to ground truth during training. The final loss function for the lane head is defined as:
\begin{equation}
 \begin{split}
    \mathcal{L}_{\textit{ls}} =\ & \lambda_{\textit{vec,ls}}\mathcal{L}_{\textit{vec}} + \lambda_{\textit{seg,ls}}\mathcal{L}_{\textit{seg}} + \lambda_{\textit{cls,ls}}\mathcal{L}_{\textit{cls}} + \lambda_{\textit{type}}\mathcal{L}_{\textit{type}} \\ & + \lambda_{\textit{top}}\mathcal{L}_{\textit{top}} + \lambda_{\textit{o2m}}\mathcal{L}_{\textit{o2m}} + \lambda_{\textit{dn}}\mathcal{L}_{\textit{dn}},
 \end{split}
\end{equation}
where 
\begin{equation}
    \mathcal{L}_{\textit{seg}} = \lambda_{ce}\mathcal{L}_{ce} + \lambda_{\textit{dice}}\mathcal{L}_{\textit{dice}}.
\end{equation}

For the road boundary head, the loss function is:
\begin{equation}
    \mathcal{L}_{\textit{rb}} = \lambda_{\textit{vec,rb}}\mathcal{L}_{\textit{vec}} + \lambda_{\textit{seg,rb}}\mathcal{L}_{\textit{seg}} + \lambda_{\textit{cls,rb}}\mathcal{L}_{\textit{cls}} + \lambda_{\textit{o2m}}\mathcal{L}_{\textit{o2m}}.
\end{equation}

The vectorized geometric loss \( \mathcal{L}_{vec} \) supervises the predicted geometry by calculating the summed Manhattan distance of the centerlines and left and right boundaries between matched lane segment pairs. 
A combined cross entropy $\mathcal{L}_{ce}$ and dice loss $\mathcal{L}_{\textit{dice}}$ is used to supervise predicted masks loss $\mathcal{L}_{\textit{seg}}$. Lane-type classification loss \( \mathcal{L}_{\textit{cls,l}} \) employs cross entropy to lane-type predictions. The topological loss \( \mathcal{L}_{\textit{top}} \) leverages focal loss \cite{lin2017iccv-flfd} to supervise the relationships among lane segments based on the topology information~\cite{li2023arxiv-gtrf}. The terms \( \mathcal{L}_{\textit{o2m}} \) and \( \mathcal{L}_{\textit{dn}} \) serve as auxiliary losses for one-to-many matching~\cite{liao2024ijcv} and denoising~\cite{li2022cvpr-dadt}, respectively.

\section{Experimental Evaluation}
\begin{table*}[tb]
  \vspace{0.1cm}
  \caption{Comparison on the OpenLane-V2 validation split (\textit{subset A}). Results for existing methods are sourced from TopoNet, with the best performance in each category highlighted in bold. $*$: baseline results taken from project page~\cite{li2024iclr}, $\dagger$: employs smaller backbone, $\ddagger$: results of the test split, CL/LS: centerline and lane segment detection respectively.}
  \label{tab:overall-results}
  \centering
  \begin{tabular}{@{}p{2.5cm}|C{0.5cm}|C{0.5cm}|C{1cm}|C{1cm}C{1cm}C{1cm}C{1cm}C{1cm}C{1cm}C{1cm}C{1cm}@{}}
    \toprule
    Method & Task & SD map & Temp. Fusion & $\text{DET}_{\textit{l/ls}}$ & $\text{DET}_{\textit{ped}}$ & $\text{DET}_{b}$ & $\text{DET}_{a}$ & $\text{DET}_{\textit{te}}$ & $\text{TOP}_{\textit{ll}}$ & $\text{TOP}_{\textit{lt}}$ & $\text{OLS}$/ $\text{OLUS}$\\
    \midrule
    MapTRv2~\cite{liao2024ijcv}      & CL  & \xmark  & \xmark &  17.7  & - & - & - & 43.5  & 5.9 & 15.2  & 31.0  \\
    TopoNet~\cite{li2023arxiv-gtrf}        & CL  & \xmark  & \xmark &28.6&-&-&-&48.6&10.9&23.8&39.8  \\
    TopoMLP~\cite{wu2024iclr}   & CL  & \xmark  & \xmark & 29.5&-&-&-&49.5&21.7&26.9&44.1  \\
    SMERF~\cite{luo2024icra}           & CL  & \cmark  & \xmark &33.4&-&-&-&48.6&15.4&23.8&42.9  \\  
    Topologic~\cite{fu2024neurips}    & CL  & \cmark  & \xmark &34.4&-&-&-&48.3&28.9&28.7&47.5  \\  
    \midrule
    LaneSegNet~\cite{li2024iclr}        & LS  & \xmark  & \xmark &  32.3  & 32.9  & - & - & -   & 25.4  & - & - \\  
    LaneSegNet$^*$~\cite{li2024iclr}    & LS  & \xmark  & \xmark &  27.8  & - & - & 23.8 & 36.9 & 24.1  & 21.3 & 36.7 \\
    TopoSD~\cite{yang2024arxiv-ttls}$^\dagger$    & LS  & \cmark  & \xmark &  37.0  & - & - & 21.6 & 40.4 & 33.6  & 24.0 & 41.2 \\    
    Score (Ours)                        & LS  & \cmark  & \cmark & \textbf{44.0}& \textbf{47.3} & \textbf{39.6} & \textbf{43.4} & \textbf{61.4} & \textbf{40.0} & \textbf{39.1} & \textbf{54.9}  \\     
    \midrule
    LaneSegNet$^\ddagger$~\cite{li2024iclr}      & LS  & \xmark & \xmark &27.3&29.2& - & -&-&22.6&-&-  \\  
    Score (Ours)$^\ddagger$              & LS  & \cmark & \cmark & \textbf{39.1}& \textbf{40.2} & \textbf{39.8} & \textbf{40.0} & \textbf{76.2} & \textbf{34.5} & \textbf{42.2} & \textbf{55.8}  \\     
    \bottomrule
  \end{tabular}
  \vspace{-0.2cm}
\end{table*}

\subsection{Datasets and Metrics}
The experiments are conducted using the \openlane\ dataset \cite{wang2023neurips}, a comprehensive dataset designed for large-scale perception and reasoning in autonomous driving scenarios. \openlane\ consists of two subsets (\textit{subset~A} and \textit{subset~B}) derived from \argo~\cite{wilson2021neurips} and \nuscenes~\cite{caesar2020cvpr}, respectively. Each subset includes 1,000 annotated scenes at a frequency of 2\,Hz.
As \textit{subset B} does not contain lane segment annotations, our work focuses on \textit{subset A}. The training set comprises roughly 27,000 frames, while the validation set includes about 4,800 frames. All lane segments within the spatial range of [\text{-}50\,m, 50\,m] along the x-axis and [\text{-}25\,m, 25\,m] along the y-axis are annotated in 3D space.
We use the \openlane\ UniScore (OLUS) as main metric:
\begin{equation}
\text{OLUS} = \frac{1}{5}[\mathrm{DET}_\textit{ls} + \mathrm{DET}_a + \mathrm{DET}_t + f(\mathrm{TOP}_{ll}) + f(\mathrm{TOP}_{lt})],
\end{equation}
where $\mathrm{DET}_\textit{ls}$, $\mathrm{DET}_a$, and $\mathrm{DET}_t$ denote the mean average precision ($\mathrm{mAP}$) for lane segment, area (road boundary and pedestrian crossing), and traffic element detection, respectively, while $\mathrm{TOP}$ refers to mAP over all vertices matched between ground truth graph and predicted graph.
For fair comparison with other methods, we explicitly report the $\mathrm{mAP}$ for pedestrian crossings $\mathrm{DET}_{\textit{ped}}$. The $\mathrm{mAP}$ for area detection is then $\mathrm{DET}_{\textit{a}} = \frac{1}{2}[\mathrm{DET}_{\textit{ped}} + \mathrm{DET}_{\textit{b}}]$ with $\mathrm{DET}_{\textit{b}}$ being the $\mathrm{mAP}$ for boundary detection.
The \openlane\ Score (OLS) is used as metric for centerline prediction and topology estimation and is reported for methods employing a centerline representation.

\openlane~\cite{wang2023neurips} provides SD map annotations. However, Luo~\etal~\cite{luo2024icra} show that leveraging OpenStreetMap (OSM) data yields better performance due to improved quality of the SD map annotations. Thus, we also leverage OSM SD maps and highlight the benefit in our ablation study.

Prior research~\cite{lilja2024cvpr, yuan2024wacv} has demonstrated that datasets utilized for HD map element detection are susceptible to data leakage, as training and validation locations exhibit substantial overlap. To assess the efficacy and resilience of our proposed approach, we conduct a comparative analysis on a geographically disjoint dataset split~\cite{yuan2024wacv}.

\subsection{Experimental Setup and Parameters}

As BEV transformation, we employ BEVFormer~\cite{li2022eccv} with a ResNet-50~\cite{he2016cvpr} backbone and a feature pyramid network (FPN)~\cite{lin2017cvpr}.
Similar to previous research~\cite{li2023arxiv-gtrf, li2024iclr}, we reduce the original image size by 50\% and leverage AdamW~\cite{kingma2015iclr} with a cosine annealing learning rate schedule for training. If not otherwise mentioned, our architecture is trained for 30 epochs with single-frame mode, followed by an additional 30 epochs dedicated to temporal fusion. For 2D detection, we train \yolo~\cite{wang2024eccv} for 200 epochs. 
The training process uses 8 NVIDIA Tesla V100 GPUs, with a total batch size of 8.

Following \cite{li2024iclr}, we select $n_{pts} = 10$ for the each polyline of the lane segment and $n_\textit{rb} = 20$ for the road boundary. All points are predicted in 3D. 
The model uses 200 base queries, supplemented by an additional $n_\textit{dn} = 60$ queries for denoising and $n_\textit{sd} = 50$ SD enhanced queries.

To solve the bipartite matching for DETR-style architectures, we use the Hungarian algorithm and follow LaneSegNet~\cite{li2024iclr} with their proposed matching costs for the lane segments. The same costs are applied for the road boundary matching. We employ the same weighting parameter $\lambda$ for both matching and loss. 
The hyperparameters for the lane head are configured as follows: $\lambda_{\textit{vec,ls}} = 0.025$, $\lambda_{\textit{seg,ls}} = 3.0$, $\lambda_{\textit{ce}} = 1.0$, $\lambda_{\textit{dice}} = 1.0$, $\lambda_{\textit{cls,ls}} = 1.5$, $\lambda_{\textit{type}} = 0.01$, $\lambda_{\textit{top}} = 5.0$, $\lambda_{\textit{o2m}} = 1.0$, and $\lambda_{\textit{dn}} = 1.0$.
The boundary head has different values for the following parameters: 
$\lambda_{\textit{vec,rb}} = 0.0125$, $\lambda_{\textit{seg,rb}} = 1.5$, and $\lambda_{\textit{cls,rb}} = 0.5$.

\subsection{Overall Results}
We benchmark our proposed method, \name, against various state-of-the-art algorithms on the OpenLane-V2 validation \textit{subset A}, as illustrated in \tabref{tab:overall-results}. 

Our approach demonstrates substantial improvements over all prior methods, attaining the best results for each task.
Notably, \name\ shows an increase of 36\% on the validation split and 43\% on the test split in lane segment detection $\text{DET}_\textit{ls}$ compared to LaneSegNet, which highlights the effectiveness of leveraging SD maps. Our method capitalizes on the structured prior knowledge provided by the SD map, allowing for superior predictions, even in challenging scenarios where parts of the lane are occluded by vehicles, as shown in \figref{fig:qualitative result}.
Further, \tabref{tab:geographicalsplit} highlights that our method also generalizes better compared to LaneSegNet due to the performance increase for the geographically disjoint dataset.
As LGMap~\cite{wu2024arxiv} and MapVision~\cite{yang2024arxiv} employ large backbones, test-time augmentation and/or ensembles, results are not directly comparable. For a comparison and further results, we would like to refer the reader to our project page.

\begin{table}[tb]
  \caption{Ablation study of individu al components. Starting from a single-frame baseline, each modification is incrementally added, demonstrating its contribution to overall performance}
  \label{tab:overall-ablation}
  \centering
  \renewcommand{\arraystretch}{1.2}
  \begin{tabular}{@{}p{3.3cm}|C{1cm}|C{1cm}|C{1cm}@{}}
    \toprule
    Method & $\text{DET}_\textit{ls}$ & $\text{DET}_{\textit{ped}}$ & $\text{TOP}_{ll}$\\
    \midrule
    Baseline                                &32.3&32.9&25.4  \\
    + Boundary \& 2D Det. Head              &32.1&34.4&28.6  \\
    + SMERF                                 &35.8&41.7&31.3  \\
    + SD Map Enhanced Queries               &37.3&41.2&31.8  \\
    + OSM SD Map                            &38.2&42.0&32.4 \\  
    + One-to-Many Matching                  &39.8&43.5&34.8 \\ 
    + Lane Denoising                        &39.8&44.1&34.6 \\ 
    + Topology Post-Processing              &39.8&44.1&36.2 \\ 
    + Dataset Resampling                    &40.0&45.1&36.2 \\ 
    + Temporal Fusion         &\textbf{44.0}&\textbf{47.3}&\textbf{40.0} \\ 
    \bottomrule
  \end{tabular}
  \vspace{0.1cm}
\end{table}
\begin{table}[tb]
  \caption{Disjoint Split Results. $*$: Epochs for temporal training.}
  \label{tab:geographicalsplit}
  \centering
  \begin{tabular}{@{}p{1.5cm}|C{1cm}|C{1cm}|C{1cm}|C{1cm}C{1cm}C{1cm}C{1cm}C{1cm}C{1cm}C{1cm}C{2cm}@{}}
    \toprule
    Method & Epochs & $DET_\textit{ls}$ & $DET_{ped}$ & $TOP_{ll}$\\
    \midrule
    LaneSegNet  &12&17.2&21.1&16.2 \\
    LaneSegNet  &24&19.3&21.6&16.9 \\
    Score (Ours)&15$^*$&\textbf{23.8}&\textbf{32.6}&\textbf{22.7} \\
    Score (Ours)&30$^*$&21.3&32.7&22.2 \\
    \bottomrule
  \end{tabular}
  \vspace{-0.2cm}
\end{table}

\begin{table}[tb]
  \vspace{0.1cm}
  \centering
  \begin{minipage}{1.5in}
    \caption{2D Traffic Element Detection Results. Traffic element heads were trained on 2D detection task only. \\$\dagger$: using YOLO proposals}
    \label{tab:2Ddetect}
    \centering
    \begin{tabular}{@{}p{2cm}|C{1cm}@{}}
      \toprule
      Method & $\text{DET}_{t}$ \\
      \midrule
      DN-DETR~\cite{li2022cvpr-dadt}  &   53.0 \\
      DN-DETR$^\dagger$  &   57.0 \\
      \yolo~\cite{wang2024eccv} &\textbf{61.4} \\
      \bottomrule
    \end{tabular}
  \end{minipage}
  \hspace{0.2in} %
  \begin{minipage}{1.5in}
    \caption{Boundary Detection Results. Detection heads were trained on the road boundary task only.\\$\dagger$: with techniques (one-to-many, temporal fusion, etc.)}
    \label{tab:boundaryresult}
    \centering
    \begin{tabular}{@{}p{2.4cm}|C{1cm}@{}}
      \toprule
      Method & $\text{DET}_{b}$ \\
      \midrule
      MapTRv2 Head~\cite{liao2024ijcv} & 31.0\\
      Lane Attention & 38.4\\
      Lane Attention$^\dagger$ & \textbf{42.5}\\
      \bottomrule
    \end{tabular}
  \end{minipage}
  \vspace{0.1cm}
\end{table}

\begin{table}[tb]
  \vspace{0.1cm}
  \caption{Results for different temporal fusion settings.}
  \label{tab:temporal}
  \centering
  \begin{tabular}{@{}C{1.1cm}|C{1.1cm}|C{1cm}|C{1cm}|C{1cm}|C{1cm}C{1cm}@{}}
    \toprule
    Epochs (single frame) & Epochs (temporal fusion) & \multirow{3}{*}{$\text{DET}_\textit{ls}$}  & \multirow{3}{*}{$\text{DET}_{\textit{ped}}$} & \multirow{3}{*}{$\text{DET}_{b}$} & \multirow{3}{*}{$\text{DET}_{\textit{ll}}$}\\
    \midrule
    30  & 0   &40&45.1&\textbf{40.5}&36.2 \\
    0   & 30   &31.1&33.6&31.8&28.4 \\
    5   & 30   &40.6&43.5&38.9&36.4\\
    30  & 30   &\textbf{44.0}&\textbf{47.3}&39.7&\textbf{40.0} \\
    \bottomrule
  \end{tabular}
  \vspace{0.2cm}
\end{table}

\subsection{Ablation Studies}
We evaluate the effectiveness of each component of \name\ in \tabref{tab:overall-ablation}. As these components primarily impact the performance of the lane segmentation head, we present results focused solely on that aspect. 
The incorporation of the SD map prior into the BEV features~\cite{luo2024icra} demonstrates an performance enhancement, while our map-enhanced queries further improve by a significant margin. The efficacy of this effect is enhanced by the utilisation of SD map data sourced from OpenStreetMap (OSM) instead of the original annotations of \openlane. Further, temporal fusion yields a substantial performance improvement.

In \tabref{tab:2Ddetect}, we present ablation studies comparing the performance of variants of DN-DETR~\cite{li2022cvpr-dadt} and \yolo~\cite{wang2024eccv}. Following~\cite{wu2024iclr}, we also evaluated DN-DETR using proposals from \yolo. For our final architecture, we chose \yolo\ due to its superior performance as 2D detection head.

To demonstrate the efficacy of our road boundary reformulation using lane attention, we present the results in \tabref{tab:boundaryresult}. 
Training on the road boundary estimation task only, our method with its reformulation outperforms a MapTRv2~\cite{liao2024ijcv} head for the same architectural design by a large margin.
Furthermore, with additional improvements such as one-to-many and temporal fusion, we achieved a 37\% increase in performance. Lastly, we analyze the impact of different warm-up durations compared to our fusion strategy on the model's performance, as shown in \tabref{tab:temporal}.

\section{CONCLUSIONS}
We propose a coherent approach for jointly predicting lane segments, road boundaries, and 2D traffic elements using onboard sensors. The network is capable of reasoning about the lane topology and associating traffic elements with lanes. It utilizes SD maps to ingest prior knowledge, which can be particularly beneficial for occluded areas as shown in the visual examples. 
The experimental results demonstrate the efficacy of the network, as it outperforms all recent models on lane segment and road boundary detection, as well as on topology reasoning. Our approach reduces the dependence on HD maps, thus, paving the way toward scalable and mapless autonomous driving.

\section{COPYRIGHT}
© 2025 IEEE.  Personal use of this material is permitted.  Permission from IEEE must be obtained for all other uses, in any current or future media, including reprinting/republishing this material for advertising or promotional purposes, creating new collective works, for resale or redistribution to servers or lists, or reuse of any copyrighted component of this work in other works.

\addtolength{\textheight}{-2cm}   %

\bibliographystyle{plain_abbrv}

\bibliography{glorified,new}

\IfFileExists{./certificate/certificate.tex}{
\subfile{./certificate/certificate.tex}
}{}

\end{document}